\documentclass[letterpaper, 10 pt, conference]{ieeeconf}  

\IEEEoverridecommandlockouts                              

\usepackage{url}
\usepackage{cite}
\usepackage{hyperref}
\usepackage{times}
\usepackage{latexsym}
\usepackage[T1]{fontenc}
\usepackage[utf8]{inputenc}
\usepackage{inconsolata}
\usepackage{graphicx}
\usepackage{booktabs}
\usepackage{graphicx}
\usepackage{pifont}
\usepackage{amssymb}
\usepackage{booktabs}
\usepackage{multirow}
\usepackage[table]{xcolor}
\usepackage{xcolor}

\title{\LARGE \bf
Can MLLMs Reason over Surround-View Driving Videos? A Multi-View Autonomous Driving QA Benchmark
}
\author{}

\author{Shunming Liu$^{1,2}$ and Jianyi Peng$^{2}$ and Cheng Li$^{2}$ and Lu Hou$^{2}$ and Pei Huang$^{3}$ and Yajing Sun$^{2,*}$ and Huaxiao Liu$^{1,*}$
\thanks{*Corresponding authors.}
\thanks{$^{1}$Shunming Liu and Huaxiao Liu are with the College of Computer Science and Technology,
        Jilin University, Changchun, China
        {\tt\small smliu23@mails.jlu.edu.cn, liuhuaxiao@jlu.edu.cn}}%
\thanks{$^{2}$Jianyi Peng, Cheng Li, Lu Hou, Yajing Sun and Shunming Liu are with Yinwang Intelligent Technology Co., Ltd., Huawei, China}%
\thanks{$^{3}$Pei Huang is with Stanford University, USA}%
}

\begin{document}

\maketitle
\thispagestyle{empty}
\pagestyle{empty}

\begin{abstract}

Surround-view video understanding in autonomous driving is a crucial capability for Multimodal Large Language Models (MLLMs) to achieve reliable driving reasoning. Real-world driving often requires information beyond the front-view camera. Existing MLLMs primarily focus on front-view information, while their spatio-temporal understanding and reasoning abilities remain unknown. To address this gap, we propose MVDriveBench, a high-information-density video QA benchmark for autonomous driving, characterized by synchronized surround-view inputs, spatio-temporal understanding, and reasoning-chain-oriented evaluation. We design 12 categories of QA tasks that cover the complete reasoning pipeline from low-level visual perception to high-level driving decision-making, enabling a systematic evaluation of MLLMs’ surround-view spatio-temporal understanding and reasoning capabilities in complex driving environments. We conduct a comprehensive evaluation of 23 open-source, closed-source, and expert MLLMs on MVDriveBench. The results show that current MLLMs still exhibit insufficient surround-view spatio-temporal reasoning ability. Meanwhile, closed-loop evaluations of specialist models in AlpaSim reveal a correlation between surround-view spatio-temporal reasoning capabilities and closed-loop driving performance, highlighting the importance of comprehensive multi-view understanding for reliable vehicle control.

\end{abstract}

\section{INTRODUCTION}

In recent years, MLLMs have made remarkable progress in image understanding, video question answering, visual reasoning, and open-ended dialogue~\cite{yin2024survey}. As their capabilities improve, systematically evaluating their perception, reasoning, and decision-making abilities has become increasingly important~\cite{fu2024mmesurvey}. Accordingly, numerous general-purpose multimodal benchmarks~\cite{li2024mvbench,fu2025videomme,wu2024longvideobench,mangalam2023egoschema,liu2024tempcompass} have been proposed, advancing MLLMs on open-domain vision-language tasks. However, these benchmarks mainly emphasize visual recognition and temporal understanding in general scenarios, and thus fail to fully capture the safety-critical, spatially constrained, and decision-oriented reasoning required for autonomous driving.

Autonomous driving is a key application for bringing MLLMs into the physical world~\cite{cui2024survey}, motivating growing interest in vision-language benchmarks for driving. However, many real-world decisions cannot rely on the front-view camera alone. For instance, lane-change safety requires jointly reasoning about vehicles across front-side, rear-side, and rear views, making many safety-critical tasks inherently dependent on synchronized multi-view videos. Existing benchmarks, however, mainly use single-view videos, discrete images, multi-view static images, or structured scene representations, and thus cannot fully evaluate MLLMs' ability to understand and reason over surround-view driving videos. This raises a fundamental question: \textit{\textbf{Do current MLLMs truly understand and reason over surround-view driving videos?}}\begin{figure}[t]
    \centering
    \includegraphics[width=0.9\linewidth]{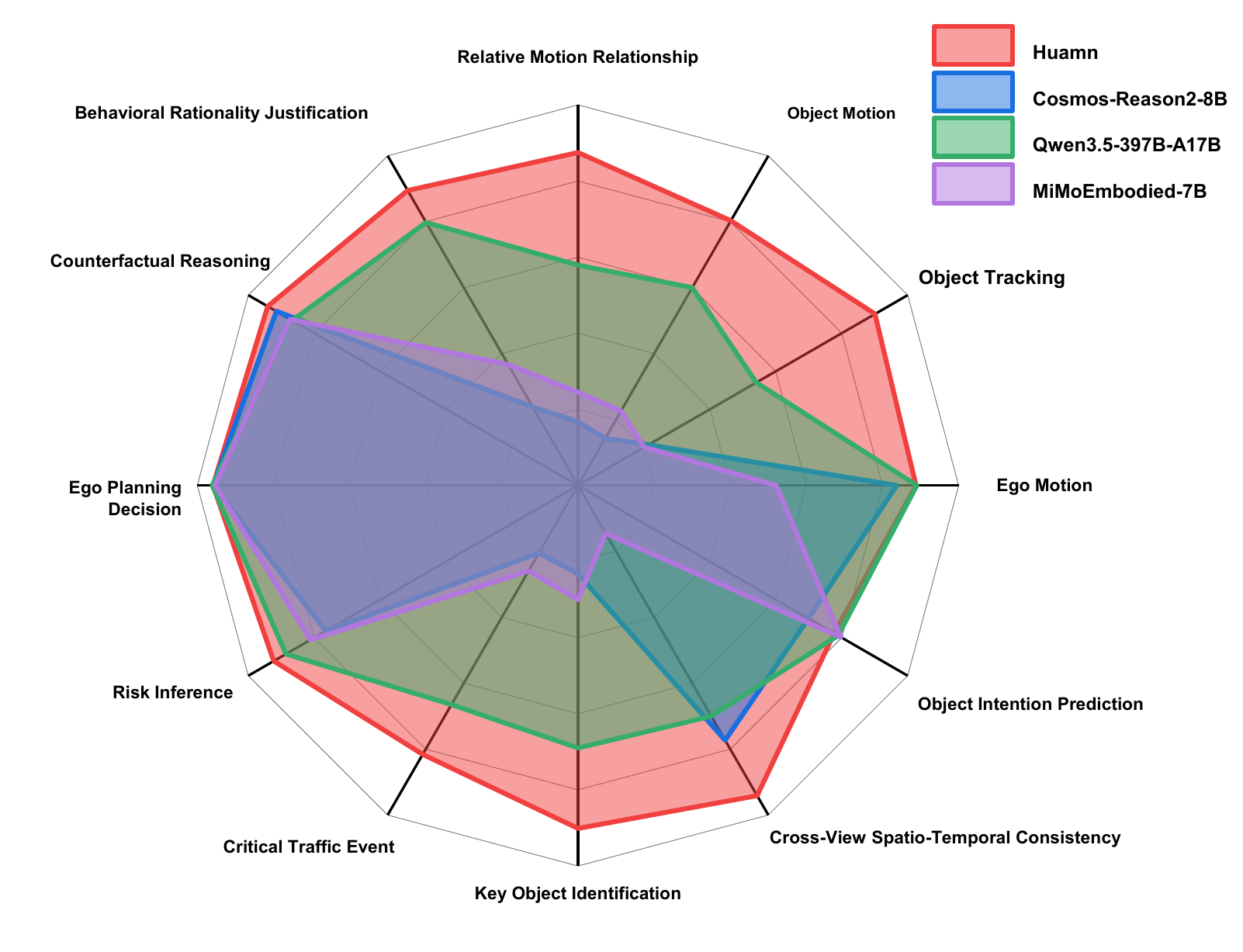}
    \caption{Performance of MLLMs on MVDriveBench.}
    \label{fig:ability_leida}
\end{figure}

To address this question, we propose \textbf{MVDriveBench}. Specifically, MVDriveBench captures the surrounding traffic environment of the ego vehicle through multiple time-synchronized onboard cameras, requiring models to understand both spatial and temporal aspects of driving scenarios. In addition, each QA sample is accompanied by a corresponding reasoning rationale as an auxiliary reference. MVDriveBench is built with a semi-automatic pipeline emphasizing multi-view necessity and two-stage prompt generation. We select real-world driving scenarios that require information beyond the front-view camera, and convert six synchronized onboard views into a unified surround-view video input. QA pairs are then generated through a two-stage process: scene understanding, which analyzes object locations, motions, cross-view relations, and risks, followed by QA generation over 12 driving reasoning categories. Manual verification is finally applied to remove unsupported or erroneous samples, yielding MVDriveBench for model evaluation.
\begin{table*}[t]
\caption{
Comparison between MVDriveBench and existing video understanding and autonomous driving visual-language benchmarks or datasets. AD denotes Autonomous Driving. Anno denotes annotation type: M = manual annotation and A = automatic annotation.
}
\centering
\footnotesize
\setlength{\tabcolsep}{3.5pt}
\renewcommand{\arraystretch}{1.15}
\resizebox{0.8\textwidth}{!}{
\begin{tabular}{lccccccccc}
\toprule
\textbf{Benchmark / Dataset} &
\textbf{AD} &
\textbf{\#Videos/Images} &
\textbf{\#QA} &
\textbf{\#Tasks} &
\textbf{Anno} &
\textbf{Closed-ended} &
\textbf{Open-ended} &
\textbf{Multi-video} &
\textbf{Multi-view} \\
\midrule
MVBench &
\ding{55} & 4K & 4K & 20 & A &
\ding{51} & \ding{55} & \ding{51} & \ding{55} \\

Video-MME &
\ding{55} & 900 & 2.7K & 30 & M &
\ding{51} & \ding{55} & \ding{51} & \ding{55} \\

LongVideoBench &
\ding{55} & 3,763 & 6,678 & 17 & M &
\ding{51} & \ding{55} & \ding{51} & \ding{55} \\

EgoSchema &
\ding{55} & 5K & 5K & 1 & M &
\ding{51} & \ding{55} & \ding{51} & \ding{55} \\

TempCompass &
\ding{55} & 410 & 7,540 & 5 & M+A &
\ding{51} & \ding{51} & \ding{51} & \ding{55} \\

\midrule
BDD-X &
\ding{51} & 6,984 & -- & 2 & M &
\ding{55} & \ding{51} & \ding{51} & \ding{55} \\

NuScenes-QA &
\ding{51} & 34K & 460K & 10 & M+A &
\ding{51} & \ding{55} & \ding{55} & \ding{51} \\

LingoQA &
\ding{51} & 28K & 419K & 4 & M &
\ding{55} & \ding{51} & \ding{51} & \ding{55} \\

DriveLM &
\ding{51} & 4871 & 443K & 5 & M+A &
\ding{51} & \ding{51} & \ding{55} & \ding{51} \\

DriveBench &
\ding{51} & 19.2K & 20,498 & 4 & M+A &
\ding{51} & \ding{51} & \ding{55} & \ding{55} \\

Reason2Drive &
\ding{51} & -- & 600K & 3 & A &
\ding{55} & \ding{51} & \ding{51} & \ding{55} \\

Talk2BEV &
\ding{51} & 1K & 20K & 4 & M &
\ding{51} & \ding{51} & \ding{55} & \ding{55} \\

DriveLMM-o1 &
\ding{51} & 539 & 4633 & 3 & M+A &
\ding{51} & \ding{51} & \ding{55} & \ding{51} \\

NuPlanQA &
\ding{51} & 90K & 1M & 9 & A &
\ding{51} & \ding{55} & \ding{55} & \ding{51} \\

STSBench &
\ding{51} & 43 & 971 & 2 & M+A &
\ding{51} & \ding{55} & \ding{51} & \ding{51} \\

\midrule
\textbf{MVDriveBench (Ours)} &
\ding{51} & \textbf{2418} & \textbf{4303} & \textbf{12} & \textbf{M+A} &
\ding{51} & \ding{51} & \ding{51} & \ding{51} \\
\bottomrule
\end{tabular}
}
\label{tab:benchmark_comparison}
\end{table*}

Overall, our evaluation of 23 MLLMs on MVDriveBench reveals that existing models still struggle with reliable surround-view driving video understanding, while Specialist Models achieve only 45.7–54.4. Meanwhile, as shown in Figure~\ref{fig:ability_leida}, the inference capability of the MoE open-source model with ultra-large parameters is also insufficient. The main contributions of this paper are summarized as follows:\begin{itemize}
    \item We propose \textbf{MVDriveBench}, a multi-view QA benchmark for surround-view video understanding in autonomous driving. It fills the gap in existing benchmarks regarding surround-view video-based driving reasoning.

    \item We systematically evaluate 23 open-source, closed-source, and expert MLLMs. The experiments reveal the capability boundaries of MLLMs in surround-view driving video understanding and reasoning from both overall accuracy and fine-grained question-type perspectives.

    \item We investigate the relationship between surround-view spatio-temporal understanding and closed-loop driving performance in AlpaSim.
\end{itemize}

\section{Related Work}
\subsection{General Video Understanding Benchmarks}
MVBench~\cite{li2024mvbench} converts public video annotations into multiple-choice QA tasks, covering 20 categories of video tasks that cannot be solved by relying on a single frame alone, thereby emphasizing temporal understanding from perception to cognition. Video-MME~\cite{fu2025videomme} further expands the coverage of general video evaluation by incorporating diverse open-domain video categories and different video lengths with manually annotated QA pairs. LongVideoBench~\cite{wu2024longvideobench} and EgoSchema~\cite{mangalam2023egoschema} respectively focus on long-context video understanding and long-horizon human activity understanding. In addition, TempCompass~\cite{liu2024tempcompass} is designed to evaluate temporal perception in videos covering temporal dimensions such as action, speed, direction, attribute change, and event order. These general video benchmarks are important for evaluating the video perception and temporal reasoning capabilities of MLLMs, but they are not specifically designed for autonomous driving scenarios.
\subsection{Vision-Language Benchmarks and Datasets for Autonomous Driving}
BDD-X~\cite{kim2018textualbddx} and LingoQA~\cite{marcu2024lingoqa} mainly rely on front-view videos, which limits their coverage of side-rear vehicles, rear-view risks, and blind spots. NuScenes-QA~\cite{qian2024nuscenesqa} constructs questions from structured annotations or scene graphs, but is closer to static scene understanding and lacks direct evaluation of synchronized video dynamics and cross-view object continuity. Talk2BEV~\cite{choudhary2024talk2bev} evaluates reasoning over BEV representations, but does not involve multi-view video reasoning. 

Recent benchmarks, including DriveLM~\cite{sima2024drivelm}, Reason2Drive~\cite{nie2024reason2drive}, and DriveLMM-o1~\cite{ishaq2025drivelmmo1}, emphasize multi-step reasoning, explanation, and planning. However, their questions do not necessarily force models to ground answers in multiple raw camera views, making it possible to rely on single-frame cues, structured representations, or language priors. DriveBench~\cite{xie2025vlmsdrivebench} focuses on visual grounding and reliability, but not on spatial-topological relationships among surround-view cameras. NuPlanQA~\cite{park2025nuplanqa} and STSBench~\cite{fruhwirth2026stsbench} are closer to multi-view driving understanding, yet they mainly focus on multi-view images, BEV-based scene understanding, or closed-form multiple-choice QA, leaving diverse surround-view video reasoning abilities insufficiently examined. 

As summarized in Table~\ref{tab:benchmark_comparison}, existing benchmarks still lack systematic evaluation of synchronized surround-view video reasoning. MVDriveBench addresses this gap by constructing QA tasks from synchronized six-view driving videos, requiring models to associate traffic participants across cameras, understand spatial continuity between adjacent views, infer temporal motion changes, and convert multi-view video evidence into spatio-temporal reasoning for safe driving.

\section{MVDriveBench}
\begin{figure*}[t]
    \centering
    \includegraphics[width=0.85\textwidth]{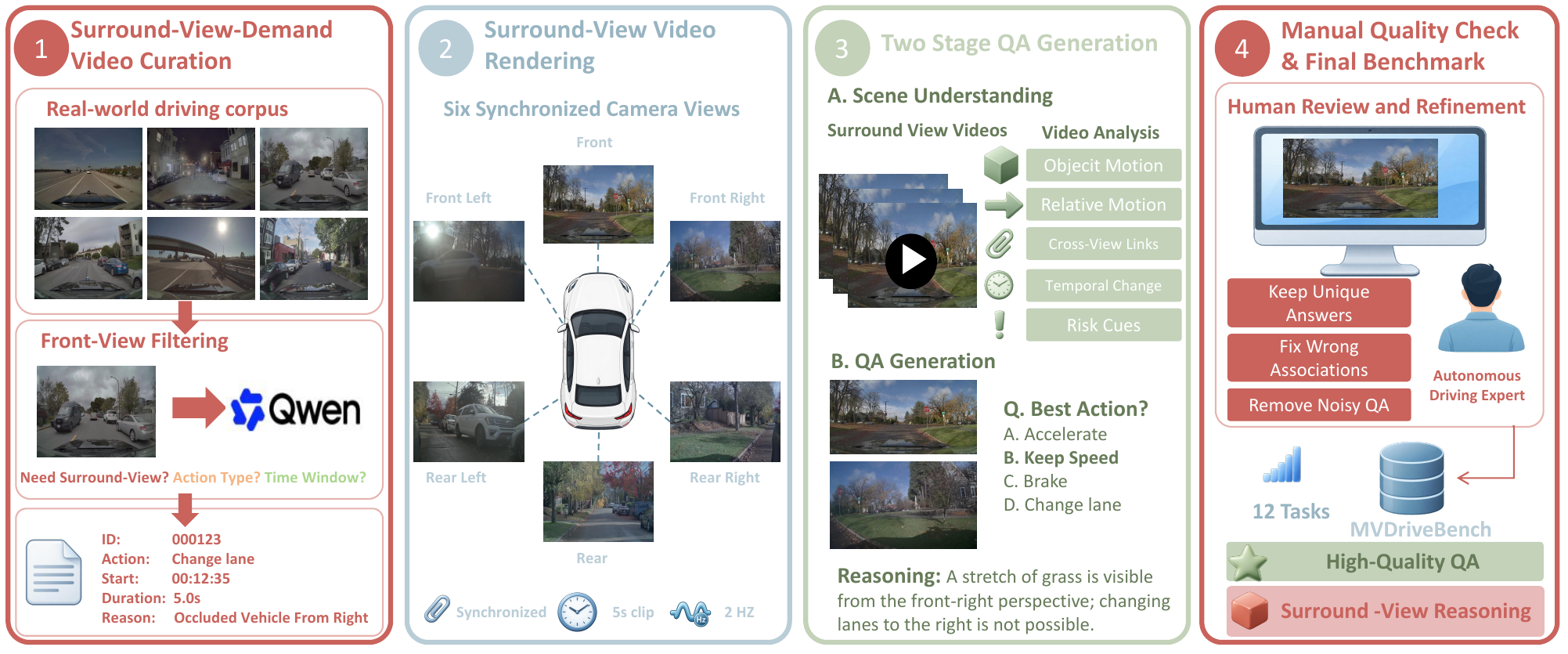}
    \caption{Illustration of the MVDriveBench annotation pipeline.}
    \label{fig:pipline_cropped}
\end{figure*}
As shown in Figure~\ref{fig:pipline_cropped}, the overall pipeline consists of four stages. Given a large-scale real-world autonomous driving video corpus $\mathcal{D}=\{c_i\}_{i=1}^{N}$, each clip $c_i$ contains synchronized multi-camera observations $\mathbf{V}_i=\{V_i^{k}\}_{k=1}^{K}$, where $K=6$ denotes the six surround-view cameras. We denote the front wide-angle camera stream as $F_i \in \mathbf{V}_i$. The goal of MVDriveBench is to construct a benchmark $\mathcal{B}$ consisting of high-quality question-answer pairs that cannot be reliably solved from $F_i$ alone, but require surround-view spatial-temporal reasoning over $\mathbf{V}_i$.
\subsection{Surround-View-Demand Video Curation}

The video clips in MVDriveBench are sourced from the NVIDIA PhysicalAI-Autonomous-Vehicles dataset~\cite{nvidia_physicalai_av_2025}. This dataset is a large-scale real-world autonomous driving dataset, covering diverse traffic flows, weather conditions, road environments, obstacles, pedestrians, and other driving-related elements.

The goal of the first stage is to select driving clips that truly require surround-view understanding. Unlike random sampling or directly using all video clips, we first determine whether each candidate clip provides sufficient spatio-temporal information for understanding the current driving state using only the front-view camera. Formally, for each clip $c_i$, only the front-view stream $F_i$ is provided to a vision-language model with a structured filtering prompt. The model predicts
\begin{equation}
g(F_i) = (m_i, a_i, s_i, d_i, r_i),
\end{equation}
where $m_i \in \{0,1\}$ indicates whether surround-view information is necessary, $a_i \in \mathcal{A}$ denotes the driving behavior type of the ego vehicle, $(s_i,d_i)$ represents the key temporal window, and $r_i$ is the filtering rationale.

Specifically, we use Qwen3.5-397B-A17B~\cite{qwen35_397b_a17b_2026} to analyze each front-view clip. The filtering prompt guides the model to reason step by step following the order of ego-vehicle motion state, surrounding traffic environment, necessity of surround-view information, and key temporal window localization. If the model determines that relying solely on the front-view perspective is insufficient to obtain the spatio-temporal information required for driving, such as observing side-rear blind spots during lane changes, assessing the motion states of vehicles in adjacent lanes during merging, or perceiving lateral oncoming traffic at intersections, the corresponding video clip is labeled as requiring surround-view understanding. The selected set is defined as
\begin{equation}
\mathcal{S} = \{c_i \in \mathcal{D} \mid m_i = 1\}.
\end{equation}
This stage also supplements each selected clip with structured metadata, including clip ID, action type, filtering reason, key start time, and duration. The output serves as the input for subsequent surround-view QA construction.

\subsection{Surround-View Video Rendering}

After obtaining candidate clips that require surround-view understanding, we construct a unified surround-view video input. For each selected clip $c_i \in \mathcal{S}$, we extract a fixed-length segment from the localized temporal window $(s_i,d_i)$ and synchronously read six onboard camera views: Front Left, Front, Front Right, Left Rear, Rear, and Right Rear. These views are rendered into a unified $2\times3$ 30 Hz grid video:
\begin{equation}
\tilde{\mathbf{V}}_i =
R\left(\{V_i^{k}[s_i:s_i+d_i]\}_{k=1}^{6}\right),
\end{equation}
where $R(\cdot)$ denotes the surround-view rendering function. This representation allows the model to observe the front, side, and rear regions around the ego vehicle within a single video input.

The six-grid representation has two advantages. First, it preserves the temporal synchronization among different cameras, enabling the model to observe the continuous appearance, disappearance, and relative motion of objects across multiple views. Second, it provides a unified input format for cross-view spatial reasoning. For example, the model can judge the boundary continuity between the Left Rear and Rear views, or combine the Front Right and Right Rear views to determine whether the right-side blind spot is occupied.
\begin{figure*}[t]
    \centering
    \includegraphics[width=0.85\textwidth]{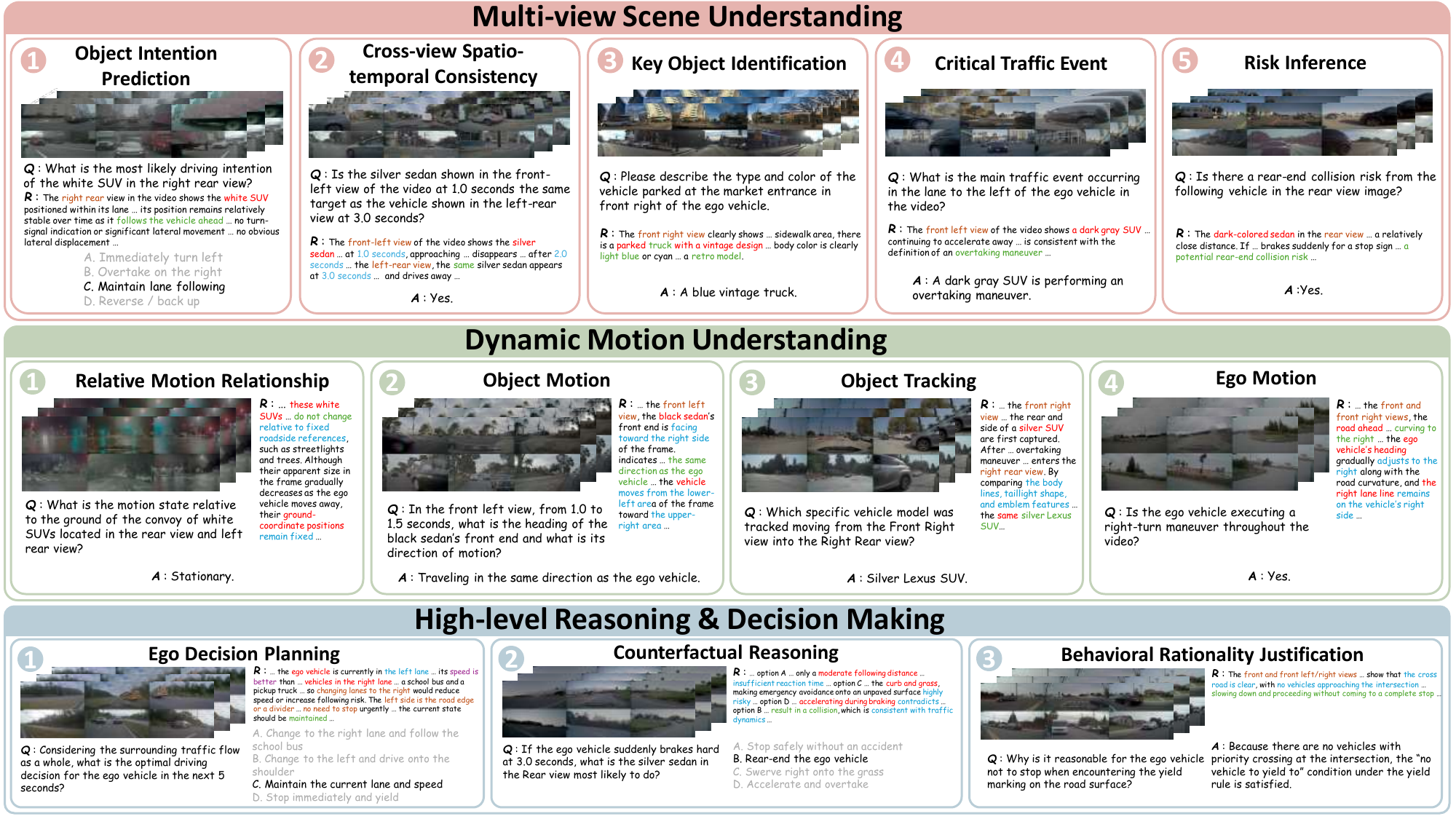}
    \caption{Illustration of the MVDriveBench's Evaluation Task.}
    \label{fig:12qa}
\end{figure*}
\begin{figure*}[t]
    \centering
    \includegraphics[width=0.85\textwidth]{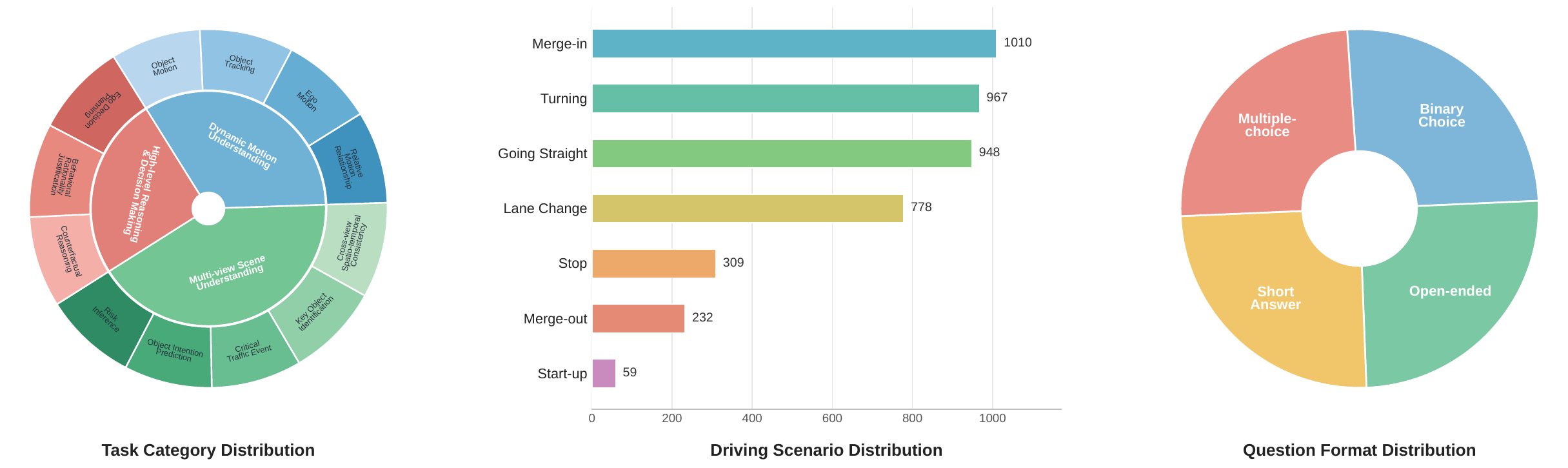}
    \caption{Statistical analysis of our MVDriveBench dataset. It consists of 3 high-level dimensions and 12 specific tasks.}
    \label{fig:benchmark_distribution_nature_vector}
\end{figure*}
\subsection{Two-Stage Prompting for Surround-View QA Generation}

Given the rendered surround-view video $\tilde{\mathbf{V}}_i$, we adopt a two-round prompting mechanism to generate high-quality QA pairs. Instead of directly asking the model to generate questions from the video, this design first requires the model to perform fine-grained scene understanding, and then generate structured QA pairs based on the intermediate understanding. This is important because surround-view driving videos contain complex spatial topology, object motion, and driving risk relationships; direct QA generation can easily lead to shallow questions or answers without sufficient visual evidence.

The first-round prompt focuses on scene understanding. The model takes the 5-second six-grid surround-view video as input and outputs a detailed scene analysis:
\begin{equation}
z_i = h_1(\tilde{\mathbf{V}}_i, p_1),
\end{equation}
where $p_1$ denotes the scene-understanding prompt, and $z_i$ contains intermediate descriptions such as object locations, object motions, relative motion relationships, cross-view associations, temporal changes, bounding-box-level evidence, and potential risks. This prompt explicitly introduces the spatial topology rules of the six views, imaging change, relative motion analysis methods and temporal analysis principle, providing an intermediate semantic foundation for subsequent QA construction.

The second-round prompt focuses on VQA generation. Within the same dialog session, the model continues to generate 12 challenging QA pairs based on the scene understanding result from the first round:
\begin{equation}
\mathcal{Q}_i = h_2(z_i, p_2)
= \{(q_{ij}, y_{ij}, \rho_{ij},\tau_{ij}, \phi_{ij})\}_{j=1}^{12},
\end{equation}
where $p_2$ denotes the QA-generation prompt, $q_{ij}$ and $y_{ij}$ represent the question and answer, $\rho_{ij}$ represents the reasoning process of the question, $\tau_{ij}\in\mathcal{T}$ denotes the question type, $\phi_{ij}\in\mathcal{F}$ denotes the question format.

The type set $\mathcal{T}$ contains 12 fine-grained driving reasoning categories in Figure~\ref{fig:12qa}, covering the full reasoning chain from low-level visual perception to high-level driving decision-making. \textit{\textbf{Relative Motion Relationships}}, \textit{\textbf{Object Motion}}, \textit{\textbf{Object Tracking}} and \textit{\textbf{Ego-vehicle Motion}} focus on dynamic changes in videos. \textit{\textbf{Intention Prediction}}, \textit{\textbf{Cross-view Spatio-Temporal Consistency}}, \textit{\textbf{Key Object Identification}}, \textit{\textbf{Critical Traffic Events}}, and \textit{\textbf{Risk Inference}} emphasize spatial topology and object continuity across multiple camera views. \textit{\textbf{Ego-vehicle Decision Planning}}, \textit{\textbf{Counterfactual Reasoning}} and \textit{\textbf{Behavior Rationality Justification}} further require the model to transform multi-view video evidence into safety-critical driving judgments.\begin{table*}[t]
\caption{Surround-view reasoning results on MVDriveBench. \textbf{Bold} indicates the best performance, while \textbf{\underline{underline}} indicates the worst performance.}
\centering
\resizebox{\textwidth}{!}{
\begin{tabular}{lcccccccccccccccc}
\toprule
\multirow{2}{*}{Model} 
& \multirow{2}{*}{Overall}
& \multicolumn{5}{c}{Dynamic Motion Understanding}
& \multicolumn{6}{c}{Multi-view Scene Understanding}
& \multicolumn{4}{c}{Driving Reasoning} \\
\cmidrule(lr){3-7} \cmidrule(lr){8-13} \cmidrule(lr){14-17}
& 
& Rel. Mot. 
& Obj. Mot. 
& Tracking 
& Ego Mot. 
& Avg
& Intent 
& Cross-view 
& Key Obj. 
& Key Event 
& Risk 
& Avg
& Planning 
& Counterfactual 
& Rationality 
& Avg \\
\midrule
Human 
& 88.5 & 87.5 & 80.3 & 90.1 & 88.8 & 86.7 
& 77.7 & 94.1 & 90.1 & 81.6 & 92.3 & 87.2 
& 95.9 & 94.1 & 89.5 & 93.2 \\

\midrule
\rowcolor{gray!15}\multicolumn{17}{l}{\textbf{\textit{Closed Source Models}}} \\
GPT-5.5~\cite{openai2026gpt55systemcard} 
& \textbf{80.1}	& 62.3	& 65.4	& 61.2	& 86.8	& 68.9	
& 85.1	& 76.7	& 73.5	& 76.9	& 92.0	& 80.8	
& 94.0	& 95.1	& 92.9	& 94.0 \\

GPT-5~\cite{singh2025openaigpt5} 
& 78.7 & 58.9 & 57.5 & 62.7 & 87.2 & 66.6
& 83.0 & 75.7 & 72.3 & 74.2 & 91.5 & 79.3
& 93.9 & 93.4 & \textbf{93.5} & 93.6 \\

GPT-4o~\cite{hurst2024gpt4o} 
& 74.3 & 48.9 & 46.3 & 50.7 & 70.4 & 54.1
& \textbf{85.3} & 89.4 & 67.9 & 70.2 & 87.8 & 80.1
& 97.0 & 92.9 & 84.6 & 91.5 \\

Gemini-2.5-Pro~\cite{comanici2025gemini25pro} 
& 76.7 & 60.8 & 56.4 & 57.1 & 87.7 & 65.5
& 68.6 & 70.5 & 74.2 & 71.6 & 92.2 & 75.4
& 95.2 & \textbf{96.7} & 88.2 & 93.4 \\

Gemini-3-Pro~\cite{gemini3pro2025} 
& 78.3 & 61.3 & 52.9 & 64.3 & 86.1 & 66.1
& 83.4 & 78.0 & 74.0 & 69.8 & \textbf{95.1} & 80.1
& 93.9 & 94.8 & 85.6 & 91.4 \\

Claude Opus 4.6~\cite{anthropic2026claudeopus46}
& 75.9	& 60.9	& 63.8	& 57.3	& 76.6	& 64.7	
& 76.6	& 80.9	& 75.8	& 70.0	& 81.7	& 77.0	
& 86.5	& 90.2	& 91.5	& 89.4 \\

\midrule
\rowcolor{gray!15}\multicolumn{17}{l}{\textbf{\textit{Open Source MoE Models}}} \\
Qwen3.5-397B-A17B~\cite{qwen35_397b_a17b_2026} 
& 74.7 & 57.9 & 60.0 & 54.1 & \textbf{89.0} & 65.2
& 78.9 & 70.0 & 69.0 & 66.5 & 88.6 & 74.6
& 96.1 & 86.7 & 79.8 & 87.5 \\

ERNIE-4.5-VL-28B-A3B-Thinking~\cite{ernie2025technicalreport} 
& 49.9 & 24.5 & 27.4 & 34.1 & 52.1 & 34.5
& 53.8 & 17.8 & 48.9 & 41.2 & 80.9 & 48.5
& 74.6 & 75.1 & 69.5 & 73.1 \\

Qwen3-VL-30B-A3B-Thinking~\cite{bai2025qwen3} 
& 66.4 & 49.0 & 58.9 & 57.7 & 52.1 & 54.4
& 80.7 & 14.6 & 70.6 & 70.1 & 81.4 & 63.5
& 92.0 & 90.1 & 81.6 & 87.9 \\

\midrule
\rowcolor{gray!15}\multicolumn{17}{l}{\textbf{\textit{Open Source Models $>$ 7B}}} \\
InternVL3-78B~\cite{zhu2025internvl3}
& 71.2 & 48.7 & 60.0 & 49.2 & 78.2 & 59.0
& 66.7 & \textbf{91.1} & 62.5 & 62.3 & 78.9 & 72.3
& 97.2 & 91.5 & 66.9 & 85.2 \\

Qwen2.5-VL-72B~\cite{bai2025qwen25vltechnicalreport} 
& 69.4 & 43.2 & 49.1 & 42.3 & 74.9 & 52.4
& 79.8 & 63.5 & 61.3 & 62.6 & 88.9 & 71.2
& 97.8 & 90.4 & 79.1 & 89.1 \\

InternVL3-38B~\cite{zhu2025internvl3} 
& 70.9 & 53.2 & 56.0 & 50.3 & 79.1 & 59.6
& 63.7 & 86.2 & 59.7 & 61.1 & 85.6 & 71.3
& 98.1 & 90.7 & 66.1 & 85.0 \\

Qwen2.5-VL-32B~\cite{bai2025qwen25vltechnicalreport} 
& 71.5 & 49.9 & 48.3 & 37.6 & 79.1 & 53.7
& 76.0 & 83.8 & 63.8 & 57.3 & 91.1 & 74.4
& 98.3 & 90.4 & 81.2 & 90.0 \\

Qwen3.5-27B~\cite{qwen35_27b_2026} 
& 61.7 & \textbf{63.2} & 61.4 & 59.3 & 52.1 & 59.0
& \textbf{\underline{26.3}} & 14.6 & \textbf{81.6} & \textbf{80.7} & 81.2 & 56.9
& 55.5 & 73.7 & 89.9 & 73.0 \\

GLM-4.1V-9B-Thinking~\cite{hong2025glm41} 
& \textbf{\underline{43.2}} & 28.1 & 34.6 & 37.4 & 54.3 & 38.6
& 32.5 & 20.5 & 54.3 & 33.8 & 81.4 & 44.5
& \textbf{\underline{29.0}} & \textbf{\underline{47.0}} & 64.2 & 46.7 \\

Video-R1-7B~\cite{feng2026videor1} 
& 62.2 & 55.2 & \textbf{73.4} & \textbf{73.4} & 69.2 & 67.8
& 51.5 & 15.1 & 24.2 & 53.0 & 74.2 & 43.6
& 93.9 & 80.5 & 73.4 & 82.6 \\

LLaVA-OneVision-Qwen2-7B-OV~\cite{li2024llavaonevision} 
& 50.5 & 26.2 & 22.9 & 33.8 & 76.0 & 39.7
& 47.1 & 35.1 & 42.1 & 35.8 & 86.1 & 49.2
& 92.0 & 80.5 & 28.1 & 66.9 \\

VGLM-QA-VGGT-8B~\cite{zheng2026learningvgllm} 
& 56.4 & 43.7 & 50.3 & 38.2 & 71.9 & 51.0
& 62.0 & 27.6 & 38.3 & 38.7 & 86.4 & 50.6
& 94.8 & 77.3 & 48.2 & 73.4 \\

\midrule
\rowcolor{gray!15}\multicolumn{17}{l}{\textbf{\textit{Specialist Models}}} \\
ReCogDrive-VLM-8B~\cite{li2026recogdrive}
& 47.3 & \textbf{\underline{14.5}} & 14.9 & \textbf{\underline{12.1}} & 73.8 & 28.8
& 69.0 & 61.1 & 18.0 & 16.6 & 86.7 & 50.3
& 97.8 & 78.8 & 24.2 & 66.9 \\

Cosmos-Reason1-7B~\cite{azzolini2025cosmosreason1} 
& 45.7 & 15.6 & \textbf{\underline{13.1}} & 15.9 & 67.8 & 28.1
& 69.6 & 82.7 & \textbf{\underline{12.2}} & \textbf{\underline{16.6}} & \textbf{\underline{51.0}} & 46.4
& \textbf{98.6} & 84.1 & \textbf{\underline{20.6}} & 67.8 \\

Cosmos-Reason2-8B~\cite{cosmos_reason2_8b_2026} 
& 51.3 & 16.7 & 14.3 & 21.4 & 83.7 & 34.0
& 69.9 & 77.3 & 23.2 & 20.5 & 76.5 & 53.5
& 95.9 & 91.5 & 23.8 & 70.4 \\

MiMoEmbodied-7B~\cite{hao2026mimoembodiedxembodiedfoundationmodel} 
& 47.3 & 24.5 & 22.6 & 20.1 & \textbf{\underline{52.1}} & 29.8
& 79.8 & \textbf{\underline{14.6}} & 30.0 & 26.0 & 81.2 & 46.3
& 95.3 & 87.3 & 36.7 & 73.1 \\

MiMo-VL-7B-SFT~\cite{coreteam2025mimovltechnicalreport} 
& 54.4 & 18.1 & 15.2 & 20.1 & 76.3 & 32.4
& 82.7 & 83.5 & 27.1 & 24.6 & 88.4 & 61.3
& 94.5 & 86.1 & 35.7 & 72.1 \\

\bottomrule
\end{tabular}
}
\label{tab:surround_reasoning_results}
\end{table*}

\begin{table*}[t]
\caption{Front-view ablation reasoning results on MVDriveBench.}
\centering
\resizebox{\textwidth}{!}{
\begin{tabular}{lcccccccccccccccc}
\toprule
\multirow{2}{*}{Model} 
& \multirow{2}{*}{Overall}
& \multicolumn{5}{c}{Dynamic Motion Understanding}
& \multicolumn{6}{c}{Multi-view Scene Understanding}
& \multicolumn{4}{c}{Driving Reasoning} \\
\cmidrule(lr){3-7} \cmidrule(lr){8-13} \cmidrule(lr){14-17}
& 
& Rel. Mot. 
& Obj. Mot. 
& Tracking 
& Ego Mot. 
& Avg
& Intent 
& Cross-view 
& Key Obj. 
& Key Event 
& Risk 
& Avg
& Planning 
& Counterfactual 
& Rationality 
& Avg \\
\midrule

\rowcolor{gray!15}
\multicolumn{17}{l}{\textbf{\textit{Closed Source Models}}} \\
GPT-5.5
& 78.8{\color{red!70!black}$\downarrow$ \scriptsize 1.3}	& 61.0	& 64.2	& 59.8	& 85.5	& 67.6	
& 82.5	& 76.5	& 73.2	& 76.5	& 89.0	& 79.5	
& 92.6	& 92.9	& 91.6	& 92.7 \\

GPT-5
& 77.2{\color{red!70!black}$\downarrow$ \scriptsize 1.5} & 64.9 & 63.5 & 54.4 & 87.9 & 67.7
& 75.0 & 63.5 & 61.7 & 78.1 & 96.6 & 75.0
& 96.5 & 90.7 & 94.5 & 93.9 \\

GPT-4o
& 65.1{\color{red!70!black}$\downarrow$ \scriptsize 9.2} & 41.5 & 35.7 & 25.0 & 77.7 & 45.0
& 77.2 & 42.7 & 54.6 & 68.6 & 85.0 & 65.6
& 98.1 & 90.7 & 85.3 & 91.4 \\

Gemini-2.5-Pro
& 73.2{\color{red!70!black}$\downarrow$ \scriptsize 3.5} & 59.4 & 49.1 & 55.2 & 84.8 & 62.1
& 66.0 & 67.1 & 61.9 & 72.0 & 87.7 & 70.9
& 96.9 & 89.8 & 87.7 & 91.5 \\

Gemini-3-Pro
& 78.4{\color{green!70!black}$\uparrow$ \scriptsize 0.1} & 50.8 & 55.2 & 65.2 & 89.6 & 65.2
& 79.2 & 87.3 & 59.2 & 75.6 & 92.3 & 78.7
& 100.0 & 94.8 & 87.8 & 94.2 \\

Claude Opus 4.6
& 73.9{\color{red!70!black}$\downarrow$ \scriptsize 2.0}	& 57.8	& 61.3	& 56.9	& 74.6	& 62.7	
& 76.3	& 77.1	& 73.4	& 66.9	& 81.3	& 75.0	
& 84.3	& 86.9	& 91.0	& 87.4 \\

\midrule
\rowcolor{gray!15}
\multicolumn{17}{l}{\textbf{\textit{Open Source MoE Models}}} \\
Qwen3.5-397B-A17B
& 59.6{\color{red!70!black}$\downarrow$ \scriptsize 15.1} & 40.1 & 46.6 & 29.4 & 75.2 & 47.8
& 73.4 & 37.0 & 33.4 & 46.6 & 89.2 & 55.9
& 90.6 & 76.8 & 77.5 & 81.6 \\

ERNIE-4.5-VL-28B-A3B-Thinking
& 43.6{\color{red!70!black}$\downarrow$ \scriptsize 6.3} & 21.4 & 26.3 & 16.5 & 52.1 & 29.1
& 43.3 & 14.9 & 33.6 & 34.4 & 80.9 & 41.4
& 70.4 & 66.3 & 63.8 & 66.8 \\

Qwen3-VL-30B-A3B-Thinking
& 53.6{\color{red!70!black}$\downarrow$ \scriptsize 12.8} & 33.1 & 31.4 & 24.5 & 52.1 & 35.3
& 71.6 & 14.3 & 35.1 & 52.2 & 81.2 & 50.9
& 92.3 & 81.3 & 75.5 & 83.0 \\

\midrule
\rowcolor{gray!15}
\multicolumn{17}{l}{\textbf{\textit{Open Source Models $>$7B}}} \\
InternVL3-78B
& 55.6{\color{red!70!black}$\downarrow$ \scriptsize 15.6} & 26.7 & 36.0 & 20.6 & 67.5 & 37.7
& 66.7 & 40.5 & 32.0 & 50.6 & 82.3 & 54.4
& 97.0 & 84.7 & 63.0 & 81.6 \\

Qwen2.5-VL-72B
& 52.4{\color{red!70!black}$\downarrow$ \scriptsize 17.0} & 20.3 & 30.3 & 15.1 & 63.9 & 32.4
& 82.2 & 16.5 & 23.2 & 44.4 & 84.8 & 50.2
& 98.3 & 84.1 & 68.1 & 83.5 \\

InternVL3-38B
& 55.9{\color{red!70!black}$\downarrow$ \scriptsize 15.0} & 34.5 & 36.0 & 21.4 & 68.0 & 40.0
& 71.3 & 25.9 & 35.0 & 48.2 & 85.3 & 53.1
& 98.6 & 89.0 & 59.1 & 82.2 \\

Qwen2.5-VL-32B
& 56.2{\color{red!70!black}$\downarrow$ \scriptsize 15.3} & 34.5 & 35.1 & 19.2 & 70.5 & 39.8
& 74.9 & 21.9 & 29.5 & 49.9 & 85.9 & 52.4
& 96.7 & 83.9 & 74.5 & 85.0 \\

Qwen3.5-27B
& 47.5{\color{red!70!black}$\downarrow$ \scriptsize 14.2} & 48.5 & 55.7 & 39.6 & 52.1 & 49.0
& 15.8 & 14.3 & 47.7 & 59.7 & 81.2 & 43.7
& 25.1 & 44.8 & 85.0 & 51.6 \\

GLM-4.1V-9B-Thinking
& 47.2{\color{green!70!black}$\uparrow$ \scriptsize 4.0} & 17.3 & 21.4 & 13.7 & 52.1 & 26.1
& 54.4 & 14.3 & 36.9 & 35.9 & 81.2 & 44.5
& 85.9 & 76.2 & 77.4 & 79.8 \\

Video-R1-7B
& 51.1{\color{red!70!black}$\downarrow$ \scriptsize 11.1} & 33.7 & 54.9 & 58.0 & 66.7 & 53.3
& 51.2 & 15.4 & 26.0 & 51.2 & 84.5 & 45.7
& 91.7 & 78.5 & 76.2 & 82.1 \\

LLaVA-OneVision-Qwen2-7B-OV
& 47.1{\color{red!70!black}$\downarrow$ \scriptsize 3.4} & 27.9 & 26.3 & 22.3 & 66.9 & 35.9
& 48.0 & 23.0 & 30.3 & 34.8 & 84.8 & 44.2
& 90.9 & 79.3 & 31.7 & 67.3 \\

VGLM-QA-VGGT-8B 
& 51.3{\color{red!70!black}$\downarrow$ \scriptsize 5.1} & 36.2 & 31.1 & 40.1 & 67.8 & 43.8
& 63.7 & 20.8 & 30.8 & 25.6 & 83.7 & 44.9
& 93.9 & 76.5 & 45.7 & 72.0 \\

\midrule
\rowcolor{gray!15}
\multicolumn{17}{l}{\textbf{\textit{Specialist Models}}} \\
ReCogDrive-VLM-8B
& 44.4{\color{red!70!black}$\downarrow$ \scriptsize 2.9} & 15.0 & 14.9 & 12.1 & 67.2 & 27.3
& 70.5 & 45.1 & 14.7 & 17.0 & 88.9 & 47.2
& 94.2 & 70.0 & 23.9 & 62.7 \\

Cosmos-Reason1-7B
& 44.1{\color{red!70!black}$\downarrow$ \scriptsize 1.6} & 10.0 & 6.3 & 12.1 & 74.1 & 25.6
& 72.8 & 58.6 & 11.1 & 13.2 & 82.0 & 47.5
& 97.0 & 77.1 & 15.7 & 63.3 \\

Cosmos-Reason2-8B
& 42.9{\color{red!70!black}$\downarrow$ \scriptsize 8.4} & 12.8 & 9.4 & 9.6 & 70.0 & 25.4
& 75.7 & 28.9 & 10.6 & 15.0 & 86.7 & 43.4
& 94.2 & 82.7 & 20.7 & 65.9 \\

MiMoEmbodied-7B
& 42.8{\color{red!70!black}$\downarrow$ \scriptsize 4.5} & 19.5 & 14.3 & 12.4 & 52.3 & 24.6
& 74.3 & 14.3 & 18.2 & 19.4 & 80.6 & 41.4
& 95.3 & 84.7 & 30.6 & 70.2 \\

MiMo-VL-7B-SFT
& 44.9{\color{red!70!black}$\downarrow$ \scriptsize 9.5} & 13.6 & 10.6 & 8.2 & 71.9 & 26.1
& 78.4 & 39.5 & 14.6 & 17.8 & 85.0 & 47.1
& 95.0 & 80.2 & 25.5 & 66.9 \\

\bottomrule
\end{tabular}
}
\label{tab:front_view_ablation_results}
\end{table*}

To prevent the final benchmark from being dominated by a few common driving behaviors, we introduce action-type-based quota control during QA generation. Let $n_a$ denote the number of generated clips for action type $a\in\mathcal{A}$, and let $b_a$ denote its target quota. We only
continue sampling clips of type $a$ when
\begin{equation}
n_a < b_a.
\end{equation}
As shown in Figure~\ref{fig:benchmark_distribution_nature_vector}, this strategy encourages a balanced distribution of different ego-vehicle behaviors.

\subsection{Manual Quality Check and Noise Filtering}

Since the QA pairs are generated with the assistance of a strong model, they still contain issues such as object recognition errors, incorrect cross-view associations, non-unique answers, or insufficient reasoning evidence. Therefore, after the generation stage, we introduce manual quality checking to remove noisy questions. Let $\mathcal{E}_i$ denote the set of manually marked invalid question indices for clip $c_i$. The final benchmark is obtained by filtering the generated QA
pairs:
\begin{equation}
\mathcal{B}
=
\bigcup_{c_i\in\mathcal{S}}
\{(q_{ij}, y_{ij}, \rho_{ij},\tau_{ij}, \phi_{ij}, a_i, c_i)
\mid j \notin \mathcal{E}_i \}.
\end{equation}

\subsection{Evaluation Protocol}

Given the constructed benchmark $\mathcal{B}$, each evaluated model $M$ produces an inference result file $\mathcal{R}_M=\{\hat{y}_n\}_{n=1}^{|\mathcal{B}|}$, where $\hat{y}_n$ denotes the model response to the $n$-th question. For each valid sample, the evaluation function assigns a score according to its question format:
\begin{equation}
s_n =
\mathrm{Eval}(q_n, y_n, \hat{y}_n, \rho_{ij}, \tau_n, \phi_n),
\end{equation}
where $y_n$ is the reference answer.

For closed-ended binary questions, we assign a binary score by exact
matching:
\begin{equation}
s_n = \mathbb{I}\left[\mathrm{Norm}(\hat{y}_n)
= \mathrm{Norm}(y_n)\right].
\end{equation}
For closed-ended multiple-choice questions, we extract the predicted option letter from the model response and compare it with the ground truth option, also yielding a $0/1$ score. For closed-ended short-answer and open-ended questions, we adopt an LLM-as-Judge protocol. \textbf{\textit{Judge Model Qwen3-8B}} in~\cite{bai2025qwen3} receives question, reference answer, model answer and the reasoning for the question, and outputs an evaluation result containing a scalar score and a brief reason:
\begin{equation}
(s_n, r_n) =
J(q_n, y_n, \hat{y}_n, \rho_{ij},\tau_n, \phi_n),
\quad s_n \in [0,1].
\end{equation}
The overall accuracy of model $M$ is computed as the mean score over all
valid samples:
\begin{equation}
\mathrm{Acc}(M)=
\frac{1}{|\mathcal{R}_M^{\mathrm{valid}}|}
\sum_{n \in \mathcal{R}_M^{\mathrm{valid}}} s_n .
\end{equation}

\section{Main Results}
Table~\ref{tab:surround_reasoning_results} shows the performance of 23 MLLMs on MVDriveBench. From the model perspective, closed-source models still maintain an overall lead, yet a clear gap remains between current MLLMs and human performance. GPT-5.5~\cite{openai2026gpt55systemcard} achieves the highest \textit{Overall} score of 80.1, but still trails human performance by 8.4 points, indicating that spatio-temporal understanding and reasoning over surround-view driving videos remain highly challenging for existing MLLMs. In contrast, Specialist Models do not exhibit the expected advantages on MVDriveBench. Driving- or embodied-oriented models such as ReCogDrive~\cite{li2026recogdrive}, Cosmos~\cite{nvidia_company_2026}, and MiMo~\cite{xiaomi_company_2026} obtain \textit{Overall} scores only in the range of 45.7–54.4, which are substantially lower than those of closed-source models and also below most large-scale open-source models. A closer inspection shows that these models generally achieve high scores on ego-vehicle decision planning tasks. For example, ReCogDrive-VLM-8B, Cosmos-Reason1-7B, and MiMoEmbodied-7B reach 97.8, 98.6, and 95.3, respectively. However, their average scores on Dynamic Motion Understanding are only 28.1–34.0, and their average scores on Multi-view Scene Understanding are merely 46.3–61.3. This suggests that their strong planning performance is not grounded in reliable surround-view perception or cross-view spatio-temporal reasoning.

\begin{figure}[t]
    \centering
    \includegraphics[width=0.85\linewidth]{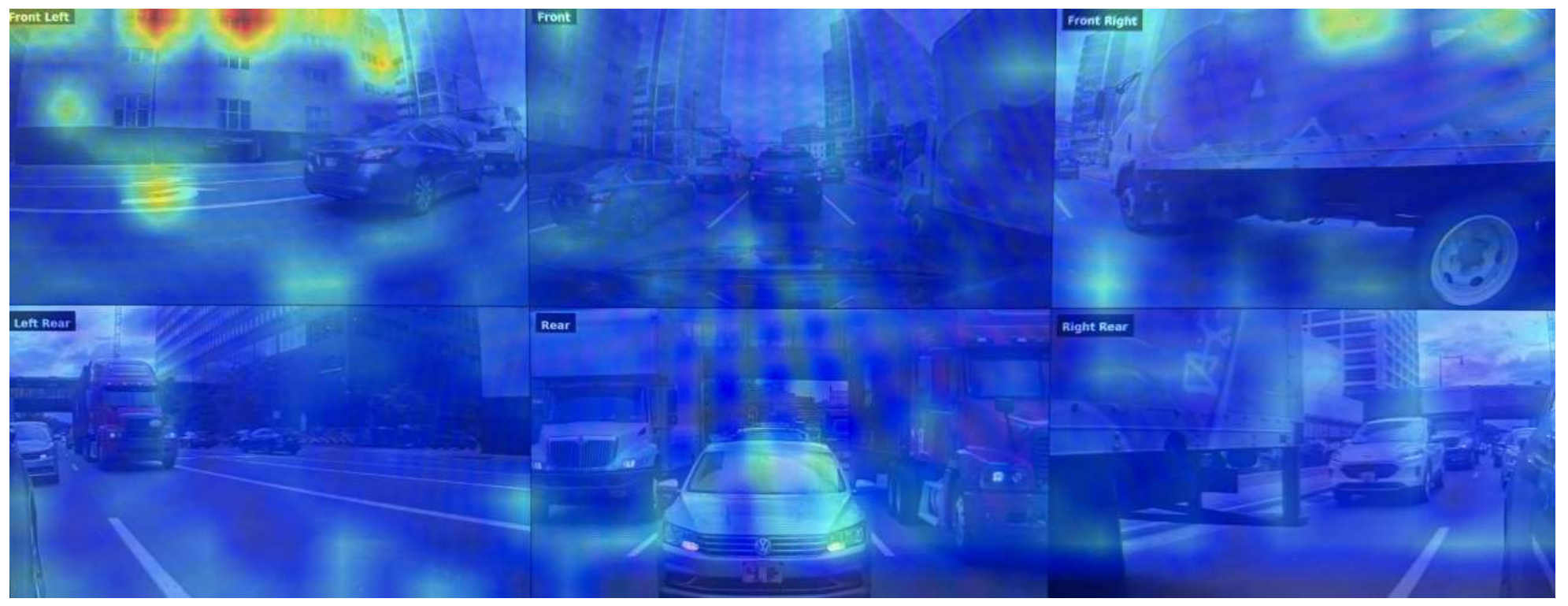}
    \caption{Heatmap of one bad case. Query: Summarize the surrounding traffic flow based on the surround-view video. Answer: Less traffic flow. GT: Heavy traffic due to a large number of vehicles.}
    \label{fig:heatmap_cropped}
\end{figure}

From the task perspective, Driving Reasoning tasks are generally the easiest for models to score highly on, but this does not imply that current models have acquired a comprehensive driving understanding. Several Specialist Models exceed 95 points on planning tasks, while their \textit{Overall} scores remain below 55, indicating that they rely more on generic driving priors rather than completing a full perception–reasoning–decision chain based on evidence from surround-view videos. By contrast, Dynamic Motion Understanding is substantially more challenging: even the best-performing closed-source model achieves an average dynamic understanding score of only 66.6, still far below the human score of 86.7. In Multi-view Scene Understanding, the gap in cross-view spatio-temporal consistency is particularly pronounced. For instance, GPT-4o~\cite{hurst2024gpt4o} and InternVL3-78B~\cite{zhu2025internvl3} achieve 89.4 and 91.1, respectively, whereas Qwen3-VL-30B-A3B-Thinking~\cite{bai2025qwen3}, Qwen3.5-27B~\cite{qwen35_27b_2026}, and MiMoEmbodied-7B~\cite{hao2026mimoembodiedxembodiedfoundationmodel} obtain only 14.6, resulting in a maximum gap of 76.5 points. Figure~\ref{fig:heatmap_cropped} visualizes that the position of Qwen3.5-27B's attention does not fall on surrounding traffic vehicles. These results indicate that cross-camera object association and multi-view spatio-temporal consistency modeling remain core bottlenecks in current surround-view driving video understanding.

\begin{figure}[t]
    \centering
    \includegraphics[width=0.85\linewidth]{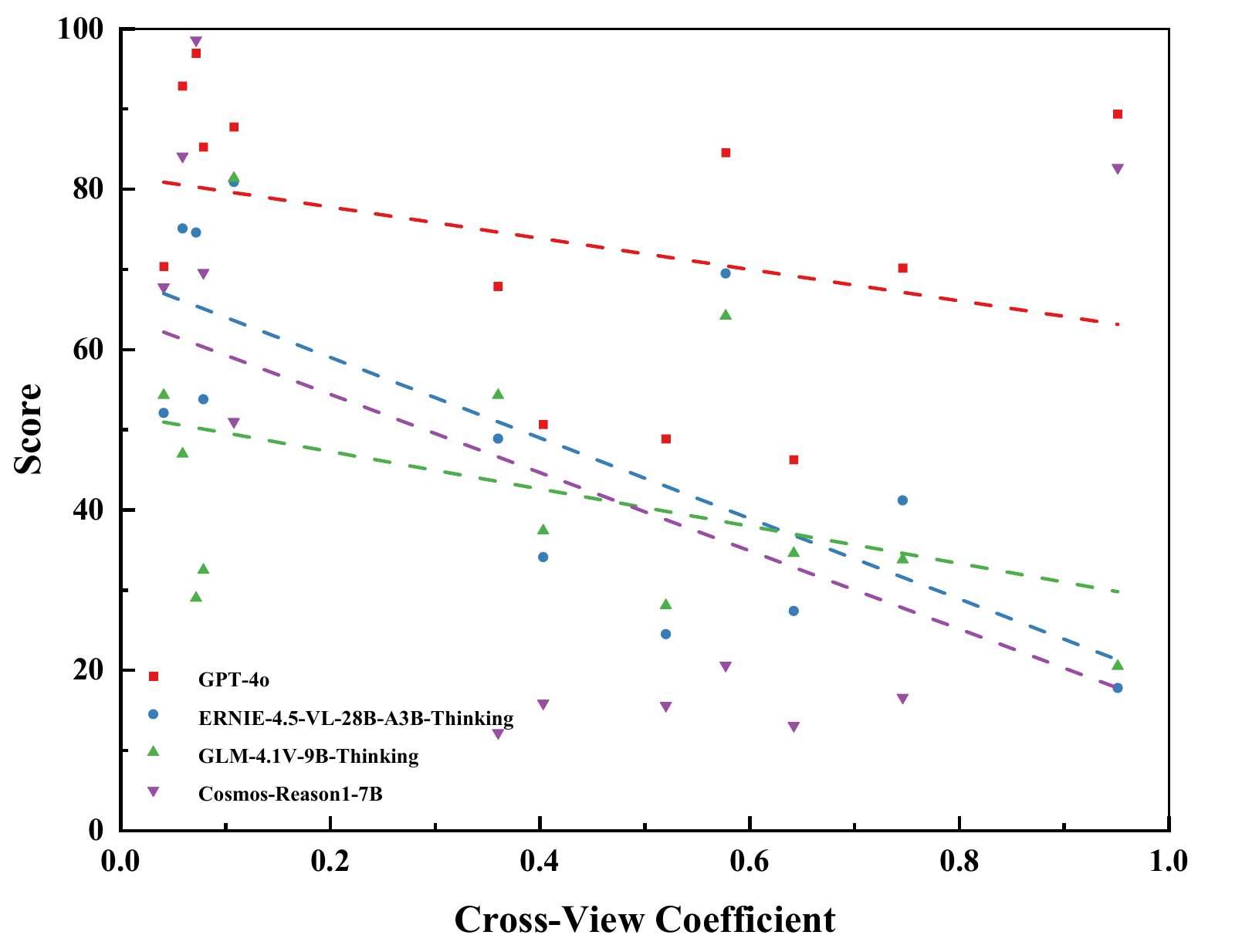}
    \caption{The trend between the proportion of multi-view questions and the response scores.}
    \label{fig:linnerfitting}
\end{figure}

As shown in Figure~\ref{fig:linnerfitting}, we compute the proportion of questions in each of the 12 task categories that require multi-view information for reliable answering, and define it as the cross-view coefficient. Through linear fitting, we observe a negative correlation between the cross-view coefficient and model performance. This indicates that current MLLMs tend to perform worse on tasks that rely more heavily on multi-view perception and reasoning.

\section{Ablation Study}
\begin{figure}[t]
    \centering
    \includegraphics[width=0.85\linewidth]{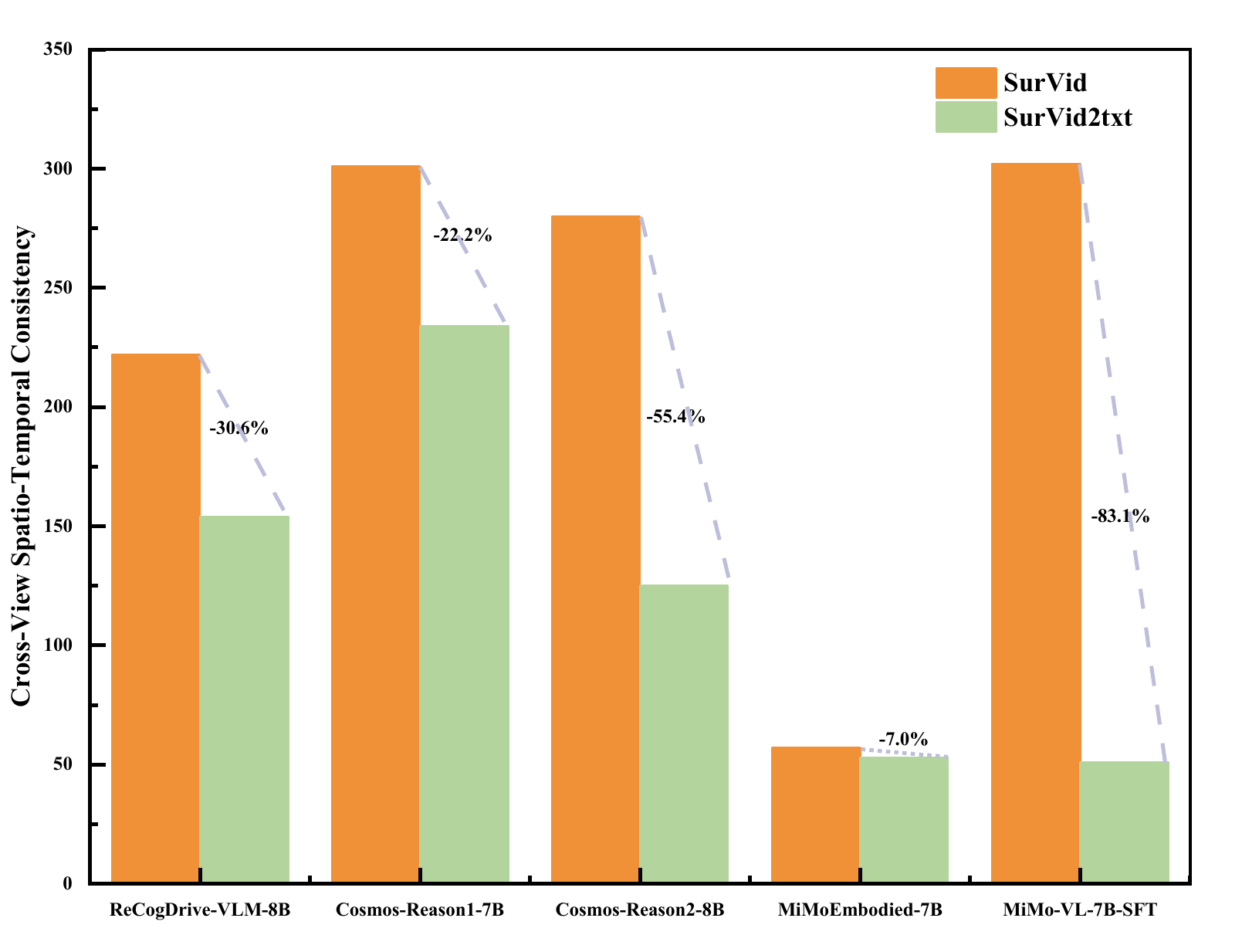}
    \caption{Performance comparison before and after converting surround-view videos into caption descriptions.}
    \label{fig:survid2txt}
\end{figure}

\begin{figure}[t]
    \centering
    \includegraphics[width=0.85\linewidth]{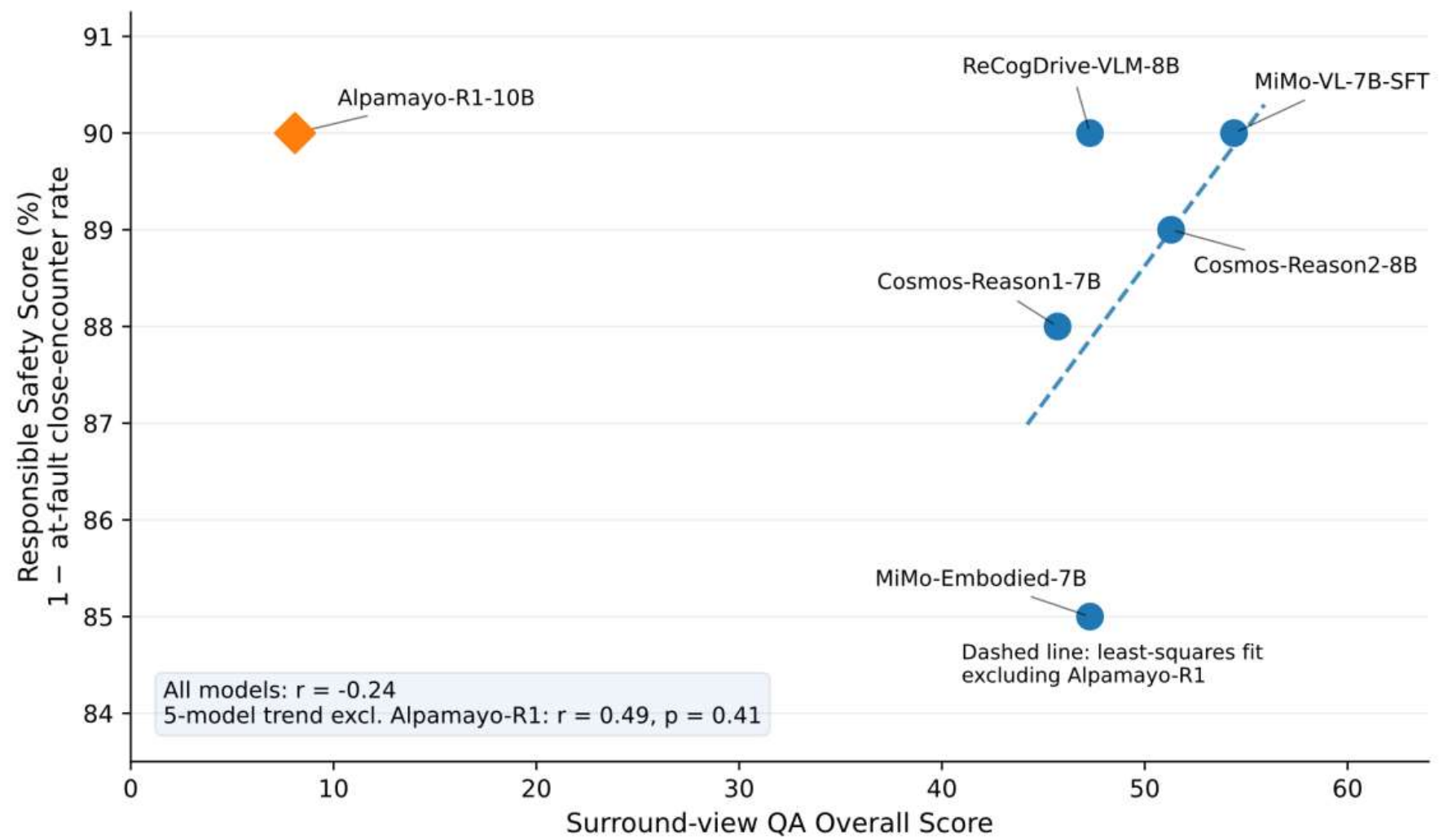}
    \caption{Relationship between surround-view QA performance and closed-loop driving safety in AlpaSim}
    \label{fig:closed_sim}
\end{figure}
\textbf{\textit{Front-View Ablation:}} As shown in Table~\ref{tab:front_view_ablation_results}, when only front-view inputs are provided, most models exhibit a substantial drop in \textit{Overall} performance. In particular, Qwen2.5-VL-72B~\cite{bai2025qwen25vltechnicalreport}, InternVL3-78B~\cite{zhu2025internvl3}, Qwen2.5-VL-32B~\cite{bai2025qwen25vltechnicalreport}, and Qwen3.5-397B-A17B~\cite{qwen35_397b_a17b_2026} decrease by 17.0, 15.6, 15.3, and 15.1 points, respectively, indicating that the front-view perspective alone is insufficient to capture the complete surrounding traffic state.

At the task level, Dynamic Motion Understanding and Multi-view Scene Understanding are the most affected categories. Tasks such as object tracking, key object identification, and cross-view spatio-temporal consistency are highly dependent on information from side and rear cameras. For example, the Cross-view score of InternVL3-78B~\cite{zhu2025internvl3} drops from 91.1 to 40.5, while the Key Object score of Qwen2.5-VL-72B~\cite{bai2025qwen25vltechnicalreport} decreases from 61.3 to 23.2. These results suggest that surround-view visual evidence is crucial for robust spatio-temporal understanding and reasoning.

Meanwhile, Driving Reasoning tasks still maintain relatively high scores under the front-view-only setting, with several models achieving planning scores above 95. This indicates that models generate plausible answers by relying on generic driving priors rather than genuinely understanding the complete surround-view scene. Overall, this ablation study shows that the question design of MVDriveBench indeed exhibits strong multi-view dependency and can effectively distinguish whether a model reasons based on real surround-view spatio-temporal evidence or merely guesses from front-view information and language priors.

\textbf{\textit{SurVid2txt Ablation:}} We convert the surround-view videos into caption descriptions and require MLLMs to answer the evaluation questions using only these textual descriptions. As shown in Figure~\ref{fig:survid2txt}, all models exhibit varying degrees of performance degradation on tasks with the highest cross-view coefficients (Cross-view Spatio-Temporal Consistency).

\section{Closed-loop Simulation} 
We further evaluate the \textit{Specialist Models} in \textbf{AlpaSim}~\cite{alpasim_2025}, a data-driven closed-loop autonomous-driving simulator that reconstructs real driving scenes and continuously renders multi-view camera observations according to the ego vehicle's predicted trajectory. Our goal is to investigate how surround-view video understanding capability influences downstream driving control, thereby complementing the open-loop QA evaluation. As shown in Figure~\ref{fig:closed_sim}, the closed-loop results show a consistent relationship with our surround-view QA benchmark. Excluding Alpamayo-R1-10B as an outlier, since its trajectory-oriented training substantially improves action prediction while degrading its general QA capability, models with higher QA scores generally achieve better closed-loop safety. For example, MiMo-VL-7B-SFT and Cosmos-Reason2-8B obtain relatively high QA scores (54.4 and 51.3) together with strong responsible safety scores (0.90 and 0.89), whereas models with weaker surround-view understanding tend to exhibit poorer closed-loop performance. These results suggest that stronger multi-view perception and reasoning abilities are generally associated with safer closed-loop driving.

\section{Conclusion}
We introduce \textbf{MVDriveBench} to evaluate MLLMs' spatio-temporal reasoning in autonomous driving. Experiments on 23 MLLMs reveal clear limitations in surround-view reasoning, while closed-loop AlpaSim results show its correlation with driving performance. We hope MVDriveBench will advance more reliable MLLMs for autonomous driving.
\vspace{-3pt}
\bibliographystyle{IEEEtran}
\bibliography{custom}

\end{document}